\pdfoutput=1

\documentclass[11pt]{article}

\usepackage[preprint]{acl}

\usepackage{times}
\usepackage{latexsym}
\usepackage{tablefootnote}

\usepackage[T1]{fontenc}

\usepackage[utf8]{inputenc}

\usepackage{microtype}
\usepackage{tcolorbox}
\usepackage{inconsolata}
\usepackage{caption}
\usepackage{subcaption}
\usepackage{graphicx}
\usepackage{multirow}
\usepackage{booktabs}
\usepackage{mathtools}
\usepackage{colortbl}
\usepackage{arydshln} 
\usepackage{amsmath} 
\usepackage{enumitem}
\usepackage{amsfonts}
\DeclareUnicodeCharacter{2212}{-}
\newcommand{\methodname}{RoLoRA}
\definecolor{mygreen}{HTML}{009901}
\definecolor{myred}{HTML}{A52A2A}

\title{RoLoRA: Fine-tuning Rotated Outlier-free LLMs for Effective Weight-Activation Quantization}

\protect
\author{Xijie Huang$^1$, Zechun Liu$^2$\thanks{All the work was done within HKUST and Zechun Liu served an advisory role.}, Shih-yang Liu$^1$, Kwang-Ting Cheng$^1$ \\     
${}^{1}$Hong Kong University of Science and Technology, ${}^{2}$Meta Reality Labs\\
\texttt{\{xhuangbs,sliuau\}@connect.ust.hk, zechunliu@meta.com,timcheng@ust.hk}}

\begin{document}
\maketitle
\begin{abstract}

Low-Rank Adaptation (LoRA), as a representative Parameter-Efficient Fine-Tuning (PEFT) method, significantly enhances the training efficiency by updating only a small portion of the weights in Large Language Models (LLMs). Recently, \textit{weight-only} quantization techniques have also been applied to LoRA methods to reduce the memory footprint of fine-tuning. However, applying \textit{weight-activation} quantization to the LoRA pipeline is under-explored, and we observe substantial performance degradation primarily due to the presence of activation outliers. 
In this work, we propose \textbf{\methodname}, the first LoRA-based scheme for effective \textit{weight-activation} quantization. {\methodname} utilizes rotation for outlier elimination and proposes rotation-aware fine-tuning to preserve the outlier-free characteristics in rotated LLMs. Experimental results show {\methodname} consistently improves low-bit LoRA convergence and post-training quantization robustness in \textit{weight-activation} settings. We evaluate {\methodname} across LLaMA2-7B/13B, LLaMA3-8B models, achieving up to 29.5\% absolute accuracy gain of 4-bit \textit{weight-activation} quantized LLaMA2-13B on commonsense reasoning tasks compared to LoRA baseline. We further demonstrate its effectiveness on Large Multimodal Models (LLaVA-1.5-7B). Codes are available at \href{https://github.com/HuangOwen/RoLoRA}{https://github.com/HuangOwen/RoLoRA}


\end{abstract}

\section{Introduction}

While we have witnessed the success of Large Language Models (LLMs) such as GPT-4~\cite{achiam2023gpt} and LLaMA~\cite{touvron2023llama} across various tasks in recent years, the massive model size and expanding training cost for LLMs have necessitated the design of model compression and Parameter-Efficient Fine-Tuning (PEFT) methods. Low-rank Adaption (LoRA)~\cite{hu2021lora}, as the most favored PEFT method, significantly enhances the fine-tuning efficiency of LLMs.

Recently, quantization techniques, which convert high-precision parameters into lower-bit formats such as INT4, have been integrated with LoRA methods~\cite{dettmers2024qlora,li2024loftq,xu2024qa,qin2024accurate}. Existing quantization-LoRA schemes can save memory costs during fine-tuning, and some schemes~\cite{li2024loftq,xu2024qa} can also reduce inference costs by producing quantized LLMs directly. However, these methods only perform \textit{weight-only} quantization, while LoRA \textit{weight-activation} quantization is under-explored. Quantizing both weights and activations in low-bit further saves run-time GPU memory and accelerates compute-intensive matrix-multiplication operations. We observe that 4-bit or 6-bit weight-activation quantization with LoRA finetuning still incurs a high accuracy degradation in LLMs, attributing to the outliers in weight and activation distribution, which stretch the quantization range and increase the quantization error.



\begin{figure}[t]
	\centering
	\includegraphics[width=\columnwidth]{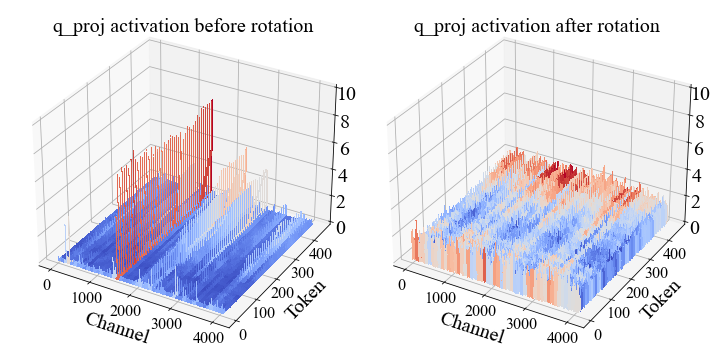}
	\caption{Activation distribution before and after rotation. The visualized input activations are selected from \textit{layers.1.self\_attn.q\_proj} in LLaMA2-7B.}
	\label{fig:activation_outlier} 
\end{figure}

Existing methods in the post-training quantization research community have endeavored to tackle the outlier challenge by mixed-precision subgrouping~\cite{zhao2024atom,chee2024quip} or shifting outliers from activation to weight~\cite{xiao2023smoothquant,shao2024omniquant}. More recently, applying rotation~\cite{ashkboos2024quarot,liu2024spinquant} to the weight matrices of LLMs has demonstrated effectiveness in eliminating activation outliers and keeping computational invariance~\cite{ashkboos2023slicegpt}. However, all these methods solve the problems from a post-training perspective, ignoring that outliers will emerge and change distribution during pre-training and fine-tuning~\cite{bondarenko2021understanding}. In this work, we take a step further to utilize the rotation for outliers-removal in LoRA fine-tuning setting and investigate the optimal solution for dynamically integrating rotation with LoRA to preserve the outlier-free characteristics and improve \text{weight-activation} quantization. Motivated by this target, we propose \textbf{R}otated \textbf{o}utlier-free \textbf{Lo}w-\textbf{R}ank \textbf{A}daptation (\textbf{\methodname}), which initially apply in-block and between-block rotation to the pre-trained LLMs, and then utilize rotation-aware fine-tuning to produce outlier-free fine-tuned LLMs as shown in Figure~\ref{fig:activation_outlier}. We explore the optimal rotation-aware fine-tuning scheme based on approximation error analysis.

Extensive experimental results prove the effectiveness of {\methodname} across diverse LLMs, tasks, and quantization settings. {\methodname} improves the 4-bit quantization for weights and activations (W4A4) performance up to 14.6 points on the MMLU benchmark compared to LoRA. Compared with existing low-bit LoRA methods, {\methodname} outperforms previous SOTA IR-QLoRA~\cite{qin2024accurate} with up to 6.0 points on the MMLU benchmark. The proposed {\methodname} is highly efficient with negligible fine-tuning overhead compared to LoRA in the same setting. {\methodname} can also improve the quantization robustness of Large Multimodal Models (LMMs) such as LLaVA~\cite{liu2024visual}, and we observe the multimodal understanding is largely retained even after W4A4 quantization as shown in Table~\ref{tab:llava_example}.

In summary, our work contributes as follows:
\begin{itemize}[itemsep=2pt,topsep=2pt,parsep=2pt]
    \item  We propose {\methodname}, the first method that explores the feasibility of integrating rotation in LoRA with quantization settings. 
    \item {\methodname} enables robust \textit{weight-activation} quantization of fine-tuned LLMs, especially in low-bit settings such as W4A4 and W6A6.
    \item We verify the effectiveness of {\methodname} on the LLaMA series (2-7B, 2-13B, 3-8B) across quantizers (RTN/GPTQ), bitwidth (W4A4/W6A6), and benchmarks (Zero-shot commonsense, MMLU). We also demonstrate that {\methodname} can be applied to LMMs.
\end{itemize}

\begin{table}[t]
\centering
\begin{minipage}{1.0\columnwidth}  
\centering
\begin{tcolorbox} 
    \centering
      \footnotesize
    \begin{tabular}{p{0.95\columnwidth}c}
\centering\includegraphics[height=4cm]{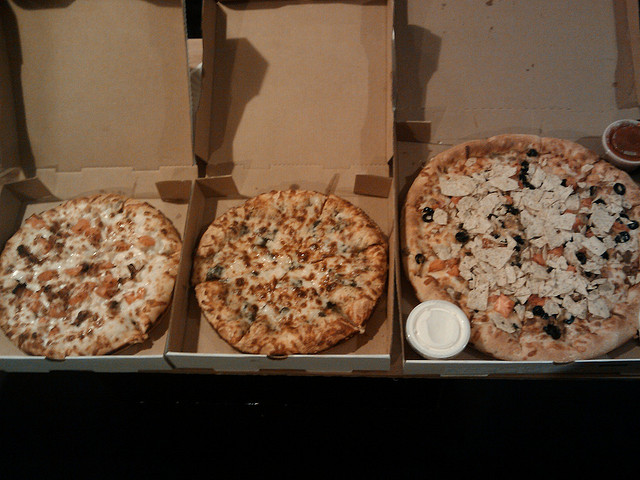} &\\
 {\bf Question:} User: Explain the visual content of the image \textcolor{blue}{in great detail}. (category: ``detail'') &\\
 \hrulefill & \\

{\bf GPT-4 Response:} The image displays three open pizza boxes with pizzas inside. The first pizza box, situated to the left, contains a smaller pizza. The second pizza box, positioned in the middle, features a larger pizza topped with tortilla chips. The third pizza box, located to the right, contains another smaller pizza. Beside the large pizza, there are two side orders of sauce in bowls, accompanying the delicious meal. The arrangement showcases a delicious spread of pizzas, inviting everyone to enjoy their different flavors and sizes.& \\
\hrulefill & \\

{\bf LoRA$\xrightarrow{\text{GPTQ}}$W4A4:} The image shows a pizza.&\\
\hrulefill & \\
{\bf \methodname$\xrightarrow{\text{GPTQ}}$W4A4:} The image features a pizza with \textcolor{blue}{a variety of toppings, including cheese, pepperonies, and chicken}. There are three pizzas in total, with one of them being a pizza with chicken. The pizzas are placed in a \textcolor{blue}{cardboard box}, which is open and ready for the pizzas. The pizzas are served in a box, making it a convenient and portable meal option.&\\
    \end{tabular}
\end{tcolorbox}
\caption{LLaVA-1.5-7B W4A4 quantization evaluation on the example from LLaVA-Bench.}
    \label{tab:llava_example}
\end{minipage} 
\end{table} 

\section{Related Work}

\noindent\textbf{Quantization} Quantization methods are powerful tools for improving training and inference efficiency. The core insight is replacing full-precision weights and activations with lower-precision representation. Most existing LLM quantization techniques fall in the category of post-training quantization (PTQ)~\citep{liu2023llm,frantar2023gptq,lin2024awq,shang2024pb,chee2024quip} that directly quantize the model without extensive training. Among these LLM PTQ methods, most of them apply \textit{weight-only} quantization while few methods explore \textit{weight-activation} quantization~\cite{xiao2023smoothquant,shao2024omniquant,zhao2024atom,ashkboos2024quarot}. Compared to the \textit{weight-only} quantization, quantizing both weights and activations enables low-precision multiply-accumulation (MAC) units. The core challenge is that outliers in activations cause high quantization errors. This work focuses on the \textit{weight-activation} quantization in the LoRA pipeline.

\noindent\textbf{LoRA} Considering that full parameter fine-tuning becomes computationally impractical as the scale of LLM continues to grow, Parameter-Efficient Fine-Tuning (PEFT) methods~\cite{li2021prefix,hu2023llm,zhang2023llama} are designed to reduce the cost by training a relatively small subset of parameters. Low-Rank Adaptation (LoRA)~\cite{hu2021lora} is the most adopted PEFT method, considering its flexibility and efficiency. More recently, LoRA variants~\cite{kopiczko2024vera,liu2024dora,hayou2024lora+} emerged to improve the effectiveness and efficiency of LoRA. Combining LoRA and quantization~\cite{dettmers2024qlora} has also been a promising direction as quantization can further save the GPU memory in LoRA finetuning. To further reduce the information distortion of low-bit finetuning, various improvements of QLoRA have been proposed~\cite{xu2024qa,li2024loftq,qin2024accurate}. 
However, these methods only apply quantization to the weight during fine-tuning to reduce memory consumption. This work is the first quantized LoRA scheme that considers the robustness to \textit{weight-activation} quantization.

\begin{figure*}[t]
	\centering
	\includegraphics[width=0.8\textwidth]{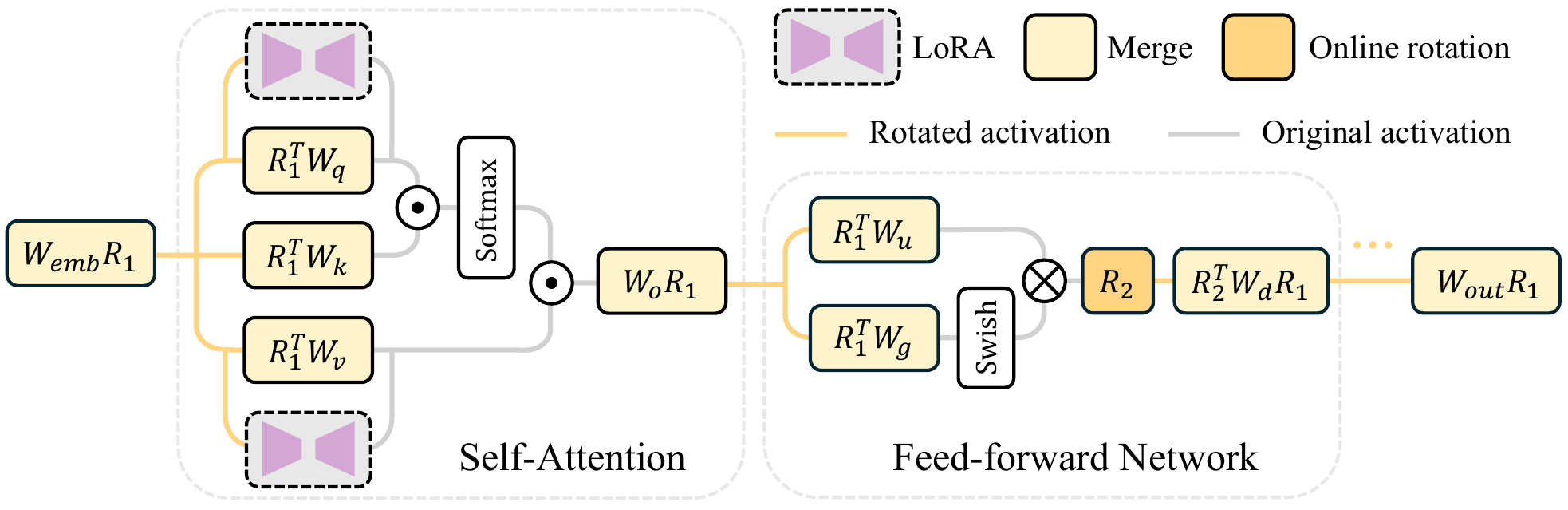}
	\caption{Overview of the proposed Rotated outlier-free LoRA (RoLoRA)}
	\label{fig:rotation_overview}
\end{figure*}

\section{Preliminary and Motivation}

\subsection{Low-Rank Adaptation (LoRA)}

For a pre-trained weight matrix $ W_0 \in \mathbb{R}^{d \times k}$, LoRA models the weight update $ \Delta W \in \mathbb{R}^{d \times k}$ utilizing a low-rank decomposition, expressed as $ AB$, where $A\in \mathbb{R}^{d \times r}$ and $ B\in \mathbb{R}^{r \times k}$ represent two low-rank matrices, with $ r \ll min(d,k)$. Consequently, the fine-tuned weight $ W'$ can be represented as:
\begin{equation} 
W' = W_0 + \Delta W = W_0 + AB,
\label{eq:lora}
\end{equation} 
where $W_0$ remains static during the fine-tuning process, and the underlined parameters are being trained. Additionally, based on Eq.~(\ref{eq:lora}), we can merge the learned $ \Delta W$ with the pre-trained weight $ W_0$ and obtain $ W'$ in advance of deployment, and given that both $ W'$ and $ W_0$ both fall within the dimensionality of $ \mathbb{R}^{d \times k}$, LoRA and its related variants do not introduce any extra latency during the inference compared to the original model. 

\subsection{Outlier in Transformer}

Starting from small-scale transformer models such as BERT and ViT, researchers have revealed that outliers exist within the weight and activation distribution~\cite{huang2023variation,wei2022outlier}. Their existence in LLMs is also observed in various studies. As shown in the left side of Figure.~\ref{fig:activation_outlier}, activation outliers are distributed per channel. While these outliers improve the representative capacity of the transformers~\cite{sun2024massive}, they bring non-trivial challenges for quantization~\cite{xiao2023smoothquant,liu2023llm}.

Most previous solutions to this outlier problem in quantization can be categorized into three types: (1) isolating these outlier values in a sub-group with higher precision, such as LLM.int8~\cite{dettmers2022gpt3}, Atom~\cite{zhao2024atom}, QuiK~\cite{ashkboos2023towards}, and AdaDim~\cite{heo2024rethinking}. However, there is non-trivial overhead for the grouping and mixed-precision. (2) shifting the challenge of quantization from activations to weights, such as SmoothQuant~\cite{xiao2023smoothquant} and OmniQuant~\cite{shao2024omniquant}. However, these methods negatively influence the weight quantization robustness and fail at W4A4 scenarios. (3) rotating activation or weight matrices to remove outliers, such as QuaRot~\cite{ashkboos2024quarot} and SpinQuant~\cite{liu2024spinquant}.  Among these methods, recent rotation-based solutions demonstrate superior effectiveness. However, previous rotation-based methods tackle the outlier challenge from a post-training perspective and have not been explored under PEFT settings. 


Thus, it leads to a question: \textit{Can we preserve the outlier-free characteristics of rotated LLMs and benefit from them during PEFT?} We show in this work that we can achieve such a target and step further to investigate the most promising rotation-based fine-tuning solutions in this work.

\subsection{Eliminating Outlier with Rotation}

A rotation matrix $ R$ is defined as an orthogonal matrix with $|R| = 1$, where $R$ also follows the characteristics of the orthogonal matrix that $ RR^\top=\mathbf{I}$. If the entries of $ R$ are either +1 or −1, it becomes a Hadamard matrix $H$. Based on the definition, we can efficiently generate $H$ with $2^k$ entries\footnote{For the $n \neq 2^k$ entries, we can also decompose it into $n = 2^km$ and construct $H_{n} = H_{m} \otimes H_{2^k}$ efficiently.} based on the Hadamard transform (also known as the Walsh–Hadamard transform~\cite{ritter1996walsh} as an example of a generalized class of Fourier transforms):
\begin{equation} 
\begin{aligned}
    H_{2^k} = \left[\begin{array}{cc}
    H_{2^{k-1}}& H_{2^{k-1}}\\ H_{2^{k-1}}& -H_{2^{k-1}}\end{array}\right]
    = H_{2} \otimes H_{2^{k-1}},
\end{aligned}
\end{equation}
where $\otimes$ denotes the Kronecker product. The rotation is highly efficient as the matrix-vector product with a $d\times d$ Hadamard matrix $H_d X$ requires $\mathcal O(d \log_2(d))$ operations. Previous research~\cite{ashkboos2023slicegpt} has revealed that applying rotation on the weights of \textit{pre-norm} transformers can retain its computational consistency and further lead to fewer outliers in the weight and activation distribution~\cite{ashkboos2024quarot,liu2024spinquant}. Concretely, the multiplication of weight matrices with a rotation matrix statistically blends weights with large and small magnitudes together into a more Gaussian-like distribution, thus producing activations with fewer outliers and easier to quantize. The outlier elimination effect of rotation is also theoretically proved in \citet{chee2024quip}.

\section{Method}

Motivated by existing challenges of activation outliers and the success of rotation-based solutions~\cite{ashkboos2024quarot,liu2024spinquant}, we introduce \textbf{R}otated \textbf{o}utlier-free \textbf{Lo}w-\textbf{R}ank \textbf{A}daptation (\textbf{\methodname}). {\methodname} initially apply in-block and between-block rotation to the pre-trained LLMs, and rotation-aware fine-tuning on the rotated LLMs will retain the optimal outlier-free characteristic, producing fine-tuned LLMs highly robust to {weight-activation} quantization.

\subsection{Applying Rotation} \label{sec:apply_rotation}
Before starting fine-tuning with rotation, we first modify the model to keep computational invariance before and after rotation. First, we need to ensure no scaling operation in the normalization module. For the LLaMA series, this can be implemented by absorbing the RMSNorm scale parameters $\alpha$ into the weight matrix right after the RMSNorm layer~\cite{elhage2023privileged}. 

Then, we perform between-block rotation to make sure that the outliers in between-block activation are eliminated. As shown in Figure~\ref{fig:rotation_overview}, we classify the weight matrices in LLMs into two groups: \textit{left-side} weights, including $ W_q, W_k, W_v$ in self-attention modules, and $W_{up}, W_{gate}$ in feed-forward network modules (which corresponds to the $W_u, W_g$ in Figure~\ref{fig:rotation_overview}). \textit{right-side} weights, including $W_o$ in self-attention modules and $ W_{down}$ in feed-forward network modules. For the weights of these two groups, we adopt different rotation strategies with
\begin{equation}
W^R_{\text{left}} \leftarrow R W_{\text{left}}, W^R_{\text{right}} \leftarrow W_{\text{right}} R^{-1},
\label{eq:weight_rotation}
\end{equation}

where the rotation $R$ is randomly generated Hadamard matrix. As we also rotated the input $X$ before embedding layer with $ X \leftarrow X R^{-1}$ and output $Y$ after \textit{lm\_head} with $Y \leftarrow RY$, the final output of the model will be identical to the original model. To avoid overflow issues in the rotation process, we converted the FP16 weights to FP64 and converted them back after the multiplication. The conversion of weight precision is only conducted once at the beginning of the rotation merging and the precision of the rotated weights will keep FP16 during the fine-tuning and inference. There will be no overhead for conversion in the actual inference because the precision during inference is always low-bit (W4A4/W6A6). These rotations are applied before any training and inference, which indicates that there will be no overhead after the merging to original weights. 

The rotation that directly applies to weights effectively reduces the outlier in between-block activation, and we refer to the operation as Between-Block Rotation (BBR). Figure.~\ref{fig:activation_outlier} demonstrates the effect of applying BBR as the activation distribution is smoother and de-centralized. However, another challenge remains that the activation in these modules still suffers from outliers, especially prevalent in FFN as discussed in previous research~\cite{bondarenko2024quantizable}. We cannot directly apply rotation similar to BBR because of the non-linear operations such as SwiGLU~\cite{shazeer2020glu} in FFN. To solve this, we adopt the online rotation node before inputting the activation input to $W_{down}$. This online rotation is implemented following the fast Hadamard kernel~\cite{chee2024quip,ashkboos2024quarot}, which can be seen as a layer dynamically rotating the activation. This online rotation operation is highly efficient as we use the fast Hadamard kernel on CUDA~\footnote{https://github.com/Dao-AILab/fast-hadamard-transform}, and the overhead is negligible during training and inference. It is referred to as In-Block Rotation (IBR). Note that IBR can also be applied to the self-attention module, but we observe in the experiments of Table~\ref{tab:rotate_where_ablation} that there is no performance improvement with this rotation.
\subsection{Rotation-aware Fine-tuning}\label{sec:raf}

\begin{figure}[t]
	\centering
	\includegraphics[width=0.95\columnwidth]{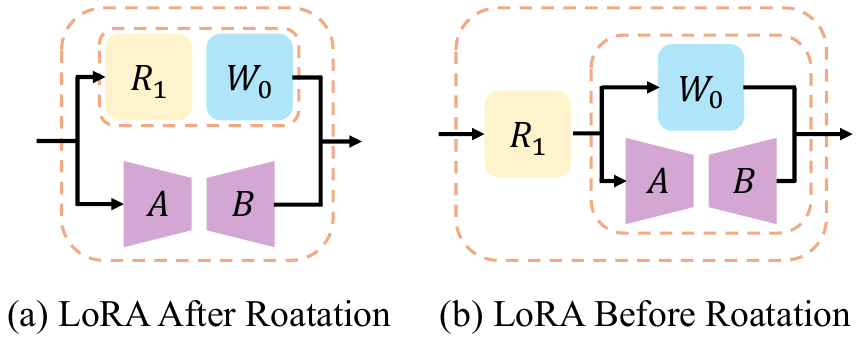}
	\caption{Two schemes for performing rotation-aware fine-tuning: (a) LAR and (b) LBR.}
	\label{fig:rotation-lora-scheme} 
\end{figure}

After performing both BBR and IBR, the between-block and in-block activation outliers are eliminated. This characteristic can lower the quantization error during QLoRA training, enabling a more accurate gradient estimation and smoother optimization for fine-tuning. However, existing research~\cite{bondarenko2021understanding,kovaleva2021bert} revealed that outliers will change distribution or emerge during fine-tuning and pre-training. This poses a new challenge of dynamically integrating rotation into LoRA to effectively maintain outlier-free characteristics. To design the optimal rotation-aware fine-tuning scheme, we first analyze the approximation difficulty when rotation is applied. We assume that the optimal weight distribution for specific downstream tasks is $W^*$, and we approximate it with the LoRA weights $AB$ merged with pre-trained weights $W_0$. The optimization of LoRA fine-tuning could be indicated as 
\begin{equation} 
\begin{aligned}
\label{eq:optimization}  
   \underset{A, B}{\min} \|W^* - (W_0 + AB)\|_{F},
\end{aligned}
\end{equation} 
where the $\|\cdot\|_{F}$ denotes the Frobenious norm. To insert the LoRA module in the rotated models, we propose two rotation-aware fine-tuning schemes, namely LoRA After Rotation (LAR) and LoRA Before Rotation (LBR), as shown in Figure~\ref{fig:rotation-lora-scheme}. 

\begin{figure}[t]
  \centering
    \includegraphics[width=0.24\textwidth]{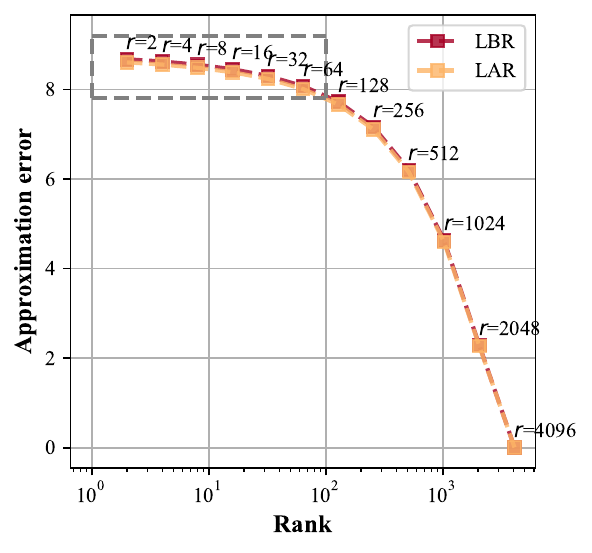}\hspace{-0.02\textwidth}
    \includegraphics[width=0.24\textwidth]{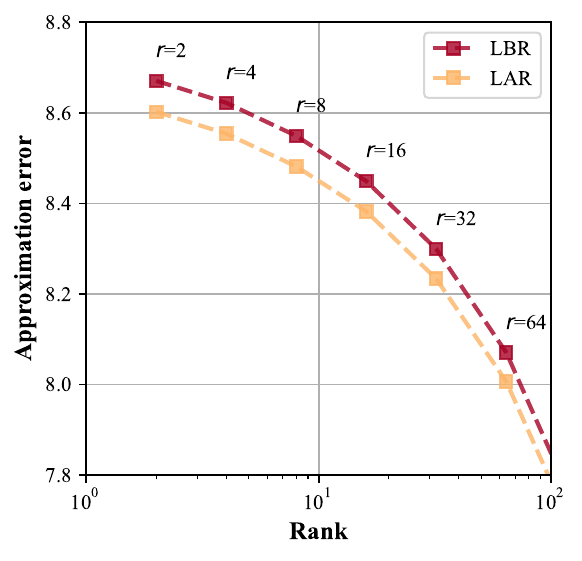}
  \caption{SVD approximation error of optimization targets with different LoRA-rotation integration schemes. }\label{fig:lora_approximation_rotate} 
\end{figure}

In LAR, we first merge the rotation matrix with pre-trained weights and then use $R_1W_0+AB$ to approximate $W^*$. For LBR, we first merge the LoRA weights and rotate them to be $R_1(W_0+AB)$. We assume the optimal weights to be the full-fine-tuning results $W_{FT}$, and the optimization for these two schemes becomes: 
\begin{equation} \small
\begin{aligned} 
\label{eq:optimization_rotate}  
   \text{LAR:} \  &\underset{A, B}{\min} \|AB-O_{\text{LAR}}\|_{F}, O_{\text{LAR}} = W_{FT}-R_1W_0\\
   \text{LBR:} \  &\underset{A, B}{\min} \|AB-O_{\text{LBR}}\|_{F}, O_{\text{LBR}} = R_1^{-1}W_{FT}-W_0
\end{aligned}  
\end{equation} 
the final optimization is very different. We apply SVD of the approximation target $O_{\text{LAR}}, O_{\text{LBR}} \in \mathbb{R}^{d \times k}$ by $O = USV^T$. The principal singular values and vectors in the first $r$ dimensions are utilized to initialize the LoRA weights with rank $r$ as $A \in \mathbb{R}^{m \times r}$ and $B \in \mathbb{R}^{r \times n}$:
\begin{equation} \small
A = U_{[:,:r]}\, S_{[:r,:r]}^{1/2} \in \mathbb{R}^{d \times r}, B = S_{[:r,:r]}^{1/2}\, V_{[:,:r]}^T \in \mathbb{R}^{r \times k}.
\end{equation}
We verify the approximation error of different rank choices $r$ to simulate the LoRA on two rotation schemes. We use a pre-trained LLaMA2-7B as $W_0$ and a full-parameter fine-tuned model on the Alpaca dataset~\cite{alpaca} as $W_{FT}$ for the experiments. which is shown in Figure.~\ref{fig:lora_approximation_rotate}. Based on the results, LAR outperforms LBR in low-rank settings with lower approximation error, suggesting LAR is the better design for rotation-aware fine-tuning. The better approximation indicates that after the two-stage merging with rotation matrices and LoRA weights, the final weights can still retain the outlier-free property, which is further validated by ablation experiments in Section~\ref{sec:ablation}.

As a result of the optimal rotation-aware fine-tuning scheme under the LAR setting, we can effectively retain the outlier-free characteristic during LLM fine-tuning, as shown in Figure~\ref{fig:kurtosis}.


\section{Experiments}

\begin{table*}[t]
\renewcommand\arraystretch{1} 
\begin{center}
\caption{Comparison of the averaged accuracy on seven Zero-shot Common Sense Reasoning (ZCSR) tasks and MMLU benchmark across LLaMA series. The detailed accuracy for each tasks are listed in Table~\ref{tab:zeroshot_full} and Table~\ref{tab:mmlu_full}.} 
\resizebox{0.95\textwidth}{!}{
\begin{tabular}{c|c|c|cc|cc|cc}
\toprule
\multirow{2}{*}{\textbf{\#Bits}} & \multirow{2}{*}{\textbf{Quantizer}} & \multirow{2}{*}{\textbf{Method}} & \multicolumn{2}{c|}{\textbf{LLaMA-2 7B}} & \multicolumn{2}{c|}{\textbf{LLaMA-2 13B}} & \multicolumn{2}{c}{\textbf{LLaMA-3 8B}} \\ 
\cmidrule{4-9}
 &  &  & ZCSR$^7$ Avg. & MMLU$^4$ Avg. & ZCSR$^7$ Avg. & MMLU$^4$ Avg. & ZCSR$^7$ Avg. & MMLU$^4$ Avg. \\
\midrule
FP16  & - & LoRA & 68.4 & 43.5 & 70.5 & 52.4 & 70.0 & 62.7 \\
\midrule
\multirow{4}{*}{W4A4}  & \multirow{2}{*}{RTN} & LoRA & 35.8 & 23.5 & 34.4 & 24.2 & 36.7 & 23.3 \\
&  & \cellcolor{gray!10}{\methodname} & \cellcolor{gray!10} \bf 54.1 (\textcolor{mygreen}{$\uparrow$18.3}) & \cellcolor{gray!10} \bf 25.8 (\textcolor{mygreen}{$\uparrow$2.3})& \cellcolor{gray!10} \bf 58.7 (\textcolor{mygreen}{$\uparrow$24.3})& \cellcolor{gray!10} \bf 30.5 (\textcolor{mygreen}{$\uparrow$6.3})& \cellcolor{gray!10} \bf 50.0 (\textcolor{mygreen}{$\uparrow$13.3})& \cellcolor{gray!10} \bf 32.1 (\textcolor{mygreen}{$\uparrow$8.8})\\
\cmidrule{2-9}
& \multirow{2}{*}{GPTQ} & LoRA & 37.0  & 23.5 & 34.4 & 24.4 & 36.6 & 23.9 \\
&  & \cellcolor{gray!10}{\methodname} & \cellcolor{gray!10} \bf 62.3 (\textcolor{mygreen}{$\uparrow$25.3})& \cellcolor{gray!10} \bf 31.0 (\textcolor{mygreen}{$\uparrow$7.5})& \cellcolor{gray!10} \bf 63.9 (\textcolor{mygreen}{$\uparrow$29.5})& \cellcolor{gray!10} \bf 38.9 (\textcolor{mygreen}{$\uparrow$14.5})& \cellcolor{gray!10} \bf 56.6 (\textcolor{mygreen}{$\uparrow$20.0})& \cellcolor{gray!10} \bf 38.5 (\textcolor{mygreen}{$\uparrow$14.6})\\
\midrule
\multirow{4}{*}{W6A6}  & \multirow{2}{*}{RTN} & LoRA & 65.3 & 35.9 & 67.3 & 47.3 & 67.7 & 55.3 \\
&  & \cellcolor{gray!10}{\methodname} & \cellcolor{gray!10} \bf 66.8 (\textcolor{mygreen}{$\uparrow$1.5})& \cellcolor{gray!10} \bf 40.5 (\textcolor{mygreen}{$\uparrow$4.6})& \cellcolor{gray!10} \bf 68.4 (\textcolor{mygreen}{$\uparrow$1.1})& \cellcolor{gray!10} \bf 47.7 (\textcolor{mygreen}{$\uparrow$0.4})& \cellcolor{gray!10} \bf 67.8 (\textcolor{mygreen}{$\uparrow$0.1})& \cellcolor{gray!10} \bf 59.4 (\textcolor{mygreen}{$\uparrow$4.1})\\
\cmidrule{2-9}
& \multirow{2}{*}{GPTQ} & LoRA & 65.5 & 35.7 & 68.0 & 47.6 & 67.8 & 54.3 \\
&  & \cellcolor{gray!10}{\methodname} & \cellcolor{gray!10} \bf 67.1 (\textcolor{mygreen}{$\uparrow$1.6})& \cellcolor{gray!10} \bf 40.8 (\textcolor{mygreen}{$\uparrow$5.1})& \cellcolor{gray!10} \bf 68.8 (\textcolor{mygreen}{$\uparrow$0.8})& \cellcolor{gray!10} \bf 47.9 (\textcolor{mygreen}{$\uparrow$0.3})& \cellcolor{gray!10} \bf 68.1 (\textcolor{mygreen}{$\uparrow$0.3})& \cellcolor{gray!10} \bf 59.4 (\textcolor{mygreen}{$\uparrow$5.1})\\
\bottomrule 
\end{tabular}}
\label{tab:main_llama}
\end{center}
\end{table*} 

\subsection{Settings}

\noindent\textbf{Model, LoRA, Quantizer} The models for our experiments include LLaMA2-7B/13B~\cite{touvron2023llama} and LLaMA3-8B~\cite{llama3modelcard}. We follow the settings in LLaMA-Factory~\cite{zheng2024llamafactory} to implement the training pipeline. The dataset for fine-tuning is Alpaca~\cite{alpaca} with 52K samples. The weight PTQ methods are the baseline Round-To-Nearest (RTN) and widely used GPTQ~\cite{frantar2023gptq}, and the activation quantizer is RTN across all experiments. We use per-channel symmetric quantization for weights and per-tensor activation quantization. 

\noindent\textbf{Tasks} Our {\methodname} was verified on seven zero-shot commonsense reasoning tasks using EleutherAI evaluation harness~\cite{gao2021framework}. These tasks include BoolQ~\citep{clark2019boolq}, PIQA~\citep{bisk2020piqa}, HellaSwag~\citep{zellers2019hellaswag}, WinoGrande~\citep{sakaguchi2021winogrande}, ARC-easy and ARC-challenge~\citep{clark2018arc}, and OBQA~\citep{mihaylov2018obqa}. Additionally, we also report the accuracy of Massively Multitask Language Understanding (MMLU) benchmark~\cite{hendrycks2020measuring} for our evaluation.

\noindent\textbf{Baselines} We consider two settings for experiments. The first is conducting FP16 fine-tuning with {\methodname}, where we compare the W4A4 and W6A6 quantization results with LoRA. The second is conducting {\methodname} fine-tuning with 4-bit weight quantization, which we refer to as Q{\methodname}, and comparing the W4A4 performance with other low-bit LoRA methods including QLoRA~\cite{dettmers2024qlora}, LoftQ~\cite{li2024loftq}, and IR-LoRA~\cite{qin2024accurate}.

\subsection{Main Results}

We first evaluate {\methodname} against LoRA in FP16 fine-tuning and then apply \textit{weight-activation} PTQ to the fine-tuned LLMs. To ensure a fair comparison, both {\methodname} and LoRA use the same settings (rank, epoch, learning rate, etc.). As listed in Table~\ref{tab:main_llama}, {\methodname} enhances the quantization robustness of the LLaMA series across various quantization settings on zero-shot commonsense reasoning and MMLU benchmarks. Specifically for the W4A4 low-bit setting, {\methodname} outperforms LoRA with an absolute up to \textbf{29.5\%} and \textbf{14.6\%} on ZCSR and MMLU, respectively. Although MMLU contains multiple-choice questions with four options. The relative accuracy below 25\% is still meaningful because we observe that some low-bit quantized LLMs cannot even be instructed to give a choice from four options. Our method can better preserve the reasoning performance, thus ensuring most of the time LLaMA is still following the instructions to answer the question rather than generating meaningless tokens. Furthermore, {\methodname} makes it feasible for near-lossless W6A6 quantization of the LLaMa series across multiple tasks.

\begin{table*}[btp]
\centering
\caption{Comparison of the averaged accuracy of different Low-bit LoRA methods on Zero-shot Common Sense Reasoning tasks and MMLU benchmark on LLaMA2-7B.} \label{tab:qlora}
\setlength{\tabcolsep}{1mm}
\resizebox{\textwidth}{!}{
\begin{tabular}{c|c|c|ccccccc:c|cccc:c}
\toprule
\textbf{\#Bits} & \textbf{Quantizer} & \textbf{Method} & \textbf{BoolQ} & \textbf{PIQA} & \textbf{HellaS.} & \textbf{WinoG.}  & \textbf{Arc-e} & \textbf{Arc-c} & \textbf{OBQA} & \textbf{Avg.}  & \textbf{Hums.} & \textbf{STEM} & \textbf{Social} & \textbf{Other} & \textbf{Avg.} \\ 
\midrule
& \multirow{4}{*}{RTN} & QLoRA~\cite{dettmers2024qlora} & 47.1 &51.5 &27.5 &49.1 &28.4 &24.6 &25.4 &36.2 &24.1 &24.7 &22.9 &21.8 &23.5 \\
&  & LoftQ~\cite{li2024loftq} & 51.5 &50.8 &26.6 &50.4 &27.5 &26.0 &25.0 & \underline{36.8} &23.9 &24.0 &22.2 &22.2 &23.2  \\
\multirow{2}{*}{W4A16}&  & IR-QLoRA~\cite{qin2024accurate} & 45.5 &49.7 &26.7 &50.6 &25.7 &26.8 &26.8 &36.0 &24.3 &24.6 &23.9 &21.9 &\underline{23.7} \\
\multirow{2}{*}{$\downarrow$}&  & \cellcolor{gray!10}{\methodname} & \cellcolor{gray!10} 59.9 &\cellcolor{gray!10} 60.5 &\cellcolor{gray!10} 43.5 &\cellcolor{gray!10} 51.8 &\cellcolor{gray!10} 43.7 &\cellcolor{gray!10} 28.6 &\cellcolor{gray!10} 28.8 &\cellcolor{gray!10} \bf 45.3 (\textcolor{mygreen}{$\uparrow$8.5})&\cellcolor{gray!10} 24.7 &\cellcolor{gray!10} 25.3 &\cellcolor{gray!10} 23.6 &\cellcolor{gray!10} 24.3 &\cellcolor{gray!10} \bf 24.5 (\textcolor{mygreen}{$\uparrow$0.8})\\
\cmidrule{2-16}
\multirow{2}{*}{W4A4}& \multirow{4}{*}{GPTQ} & QLoRA~\cite{dettmers2024qlora} & 51.4 &51.6 &27.7 &51.9 &29.6 &25.3 &26.4 & \underline{37.7} &24.9 &24.0 &22.2 &22.5 &23.6  \\
&  & LoftQ~\cite{li2024loftq} & 55.9 &49.2 &27.2 &49.1 &26.6 &26.1 &24 &36.9 &24.1 &23.8 &23.3 &22.7 &23.6  \\
&  & IR-QLoRA~\cite{qin2024accurate} & 51.1 &49.8 &27.6 &49.3 &27.6 &24.6 &27.4 &36.8 &24.6 &24.8 &22.9 &22.7 & \underline{23.9} \\
&  & \cellcolor{gray!10}{\methodname} & \cellcolor{gray!10} 68.7 &\cellcolor{gray!10} 73.1 &\cellcolor{gray!10} 66.8 &\cellcolor{gray!10} 61.3 &\cellcolor{gray!10} 61.2 &\cellcolor{gray!10} 37.8 &\cellcolor{gray!10} 38.2 &\cellcolor{gray!10} \bf 58.2 (\textcolor{mygreen}{$\uparrow$20.5}) &\cellcolor{gray!10} 28.3 &\cellcolor{gray!10} 32.7 &\cellcolor{gray!10} 32.3 &\cellcolor{gray!10} 27.2 &\cellcolor{gray!10} \bf 29.9 (\textcolor{mygreen}{$\uparrow$6.0})\\
\bottomrule
\end{tabular}}  
\end{table*}    

We further evaluate {\methodname} against QLoRA~\cite{dettmers2024qlora} and serval baseline methods, including LoftQ~\cite{li2024loftq}, IR-QLoRA~\cite{qin2024accurate}, on 4-bit fine-tuning and then apply W4A4 PTQ to the low-bit fine-tuned LLaMA2-7B. The performance across seven commonsense reasoning tasks and four MMLU subtasks is detailed in Table~\ref{tab:qlora}. We can see that {\methodname} consistently improves the performance of the quantized model using the same quantizer. In particular, for W4A4 GPTQ,  {\methodname} exceeds QLoRA by \textbf{20.5\%} on the average accuracy of commonsense reasoning tasks. Across the experiments on both FP16 and 4-bit fine-tuning, we observe that {\methodname} achieves higher performance improvement on the LLMs quantized by GPTQ~\cite{frantar2023gptq} in general. This observation supports our claim that {\methodname} retains the outlier-free activation in fine-tuning as GPTQ only helps lower the quantization error of weights but not for activation.

\subsection{Visual Instruction Tuning}

We further verify the effectiveness of {\methodname} on visual instruction tuning tasks with LLaVA-1.5-7B~\cite{liu2023improved}, which consists of a language model, Vicuna-7B~\cite{vicuna2023}, and a vision encoder CLIP ViT-L-336px~\cite{radford2021learning}. We finetune the LLaVA-1.5-7B on LLaVA-Instruct-150K\footnote{https://huggingface.co/datasets/liuhaotian/LLaVA-Instruct-150K}. We only perform quantization on the language model and evaluate the LLaVA with quantized Vicuna and full-precision vision encoder on LLaVA-bench (COCO)~\cite{liu2024visual} with GPT-4~\cite{achiam2023gpt}. The relative score across the conversation, detail description, and complex reasoning are reported in Table.~\ref{tab:llava_bench}, where we can observe from the results that {\methodname} help improve the quantization robustness and keep the multi-modal ability during PTQ to the better extent with an increase up to 18.9 overall scores. We also provide an example of the detail description task on a given image shown in Table.~\ref{tab:llava_example}. While the W4A4 LoRA model only gives a rough superficial description of the images, our W4A4 {\methodname} model fully elaborates the details, such as the toppings and containers.

\begin{table}[h] 
\caption{Comparison of the W4A4 quantization performance on LLaVA-Bench of LLaVA-1.5-7B.}\label{tab:llava_bench}
\centering
\resizebox{0.48\textwidth}{!}{\begin{tabular}{c|c|ccccc}
\toprule
\textbf{\#Bits} & \textbf{Quantizer} & \textbf{Method} & \textbf{Conv.} & \textbf{Detail} & \textbf{Reas.} & \textbf{Overall}  \\ 
\midrule
\multirow{4}{*}{W4A4}  & \multirow{2}{*}{RTN} & LoRA &  43.2 & 29.6 & 31.6 & 34.9 \\
&  & \cellcolor{gray!10}{\methodname} & \cellcolor{gray!10} 68.8 & \cellcolor{gray!10} 40.5 & \cellcolor{gray!10} 51.9 & \bf \cellcolor{gray!10} 53.8 (\textcolor{mygreen}{$\uparrow$18.9})\\
\cmidrule{2-7}
& \multirow{2}{*}{GPTQ} & LoRA & 70.6 & 41.8 & 47.9 & 53.5 \\
&  & \cellcolor{gray!10}{\methodname} & \cellcolor{gray!10} 67.5 & \cellcolor{gray!10} 48.3 & \cellcolor{gray!10} 66.2 & \cellcolor{gray!10} \bf 60.8 (\textcolor{mygreen}{$\uparrow$7.3}) \\
\bottomrule
\end{tabular}}  
\end{table}

\subsection{Compatibility with other LoRA variants}

We further verify our method on a representative LoRA variant, DoRA~\cite{liu2024dora}. DoRA decomposes the pre-trained weight into magnitude and directional components and finetunes both. We also follow this scheme in our rotation-aware fine-tuning stage and refer to this scheme as RoDoRA. As shown in Table~\ref{tab:dora}, RoDoRA achieves 18.3\% and 26.7\% higher accuracy on W4A4 LLaMA2-7B using RTN and GPTQ as quantizers. The results of RoDoRA also outperform RoLoRA, showing the compatibility of our methods with cutting-edge LoRA variants and potential to further enhance the performance of weight-activation quantization.

\begin{table}[h]  
\caption{Compatibility of with DoRA on LLaMA2-7B.}\label{tab:dora}
\centering
\resizebox{0.45\textwidth}{!}{\begin{tabular}{c|c|cc}
\toprule
\textbf{\#Bits} & \textbf{Quantizer} & \textbf{Method} & \textbf{ZCSR$^7$ Avg.}  \\ 
\midrule
\multirow{4}{*}{W4A4}  & \multirow{2}{*}{RTN} & DoRA~\cite{liu2024dora} &  36.4  \\
&  & \cellcolor{gray!10} RoDoRA & \cellcolor{gray!10} \bf 54.7 (\textcolor{mygreen}{$\uparrow$18.3})  \\
\cmidrule{2-4}
& \multirow{2}{*}{GPTQ} & DoRA~\cite{liu2024dora} & 36.6  \\
&  & \cellcolor{gray!10} RoDoRA & \cellcolor{gray!10} \bf 63.3 (\textcolor{mygreen}{$\uparrow$26.7}) \\
\bottomrule
\end{tabular}}
\end{table}  

\begin{figure*}[h]
  \centering
    \includegraphics[width=0.66\columnwidth]{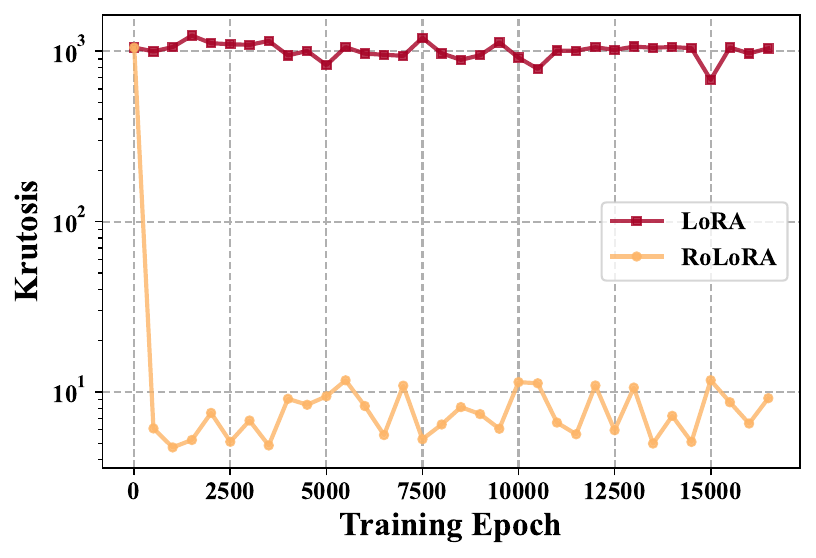} 
    \includegraphics[width=0.67\columnwidth]{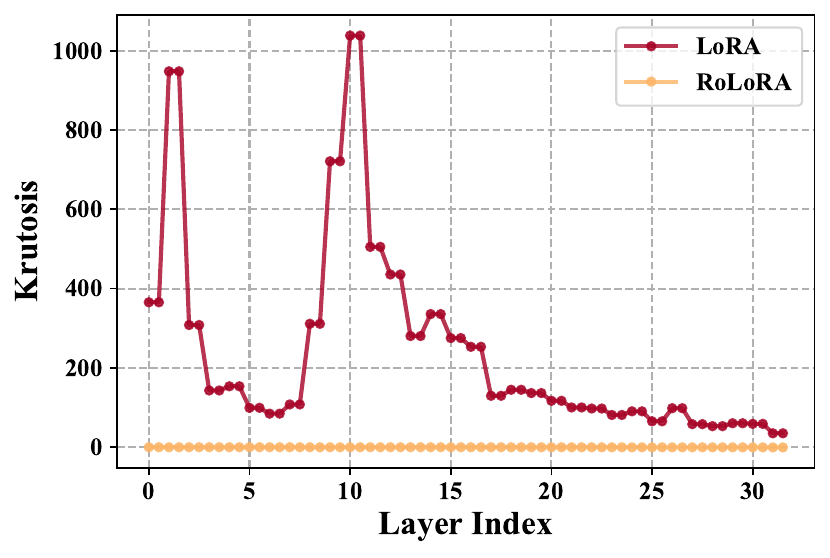}
    \includegraphics[width=0.65\columnwidth]{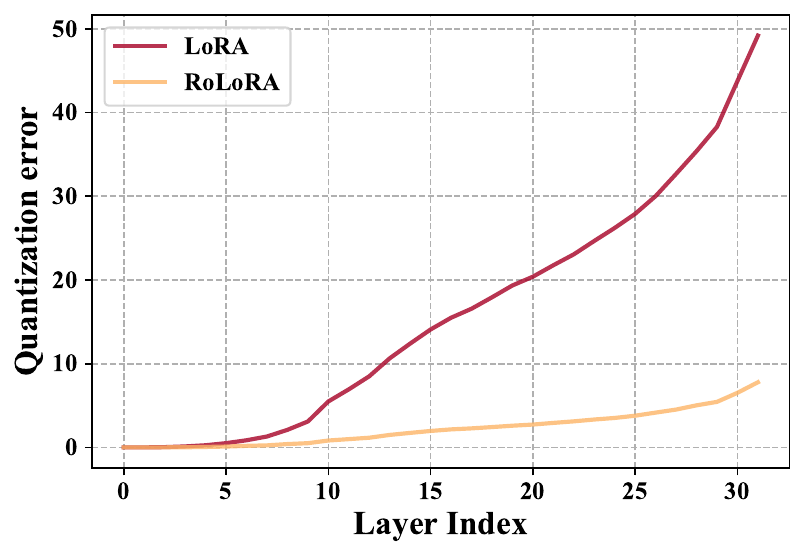}
  \caption{\textbf{Left:} The training dynamics of the average Kurtosis of activations, \textbf{Middle:} The distribution of Kurtosis of activations across all layers in the final model after fine-tuning with LoRA and {\methodname}, \textbf{Right:} The accumulative quantization error of W4A4 GPTQ across all layers in the final model after fine-tuning with LoRA and {\methodname}.}
  \label{fig:kurtosis}
\end{figure*}

\subsection{Ablation Study and Analysis}\label{sec:ablation}

\noindent\textbf{\underline{When} to Apply Rotation?} Different from the Rotation-Aware Fine-tuning (RAF) scheme that rotates the LLMs before LoRA fine-tuning, we can also directly apply rotation on an already-finetuned LoRA model. This possible paradigm of LoRA$\rightarrow$Rotate$\rightarrow$PTQ is referred to as post-training rotation. We evaluate post-training rotation using the same training setting as {\methodname} across the LLaMA series. The W4A4 GPTQ performance on seven zero-shot commonsense reasoning tasks are listed in Table~\ref{tab:when_rotate}. The results indicate that applying rotation before LoRA can consistently enhance the quantization robustness of the fine-tuned LLMs.

\begin{table}[h]   
	\caption{Ablation on \textbf{when} to apply rotation.}
	\label{tab:when_rotate}
	\centering
 \setlength{\tabcolsep}{0.5mm}
	\resizebox{0.48\textwidth}{!}{
		\begin{tabular}{cccc}
			\toprule
			\textbf{Method}  & \textbf{LLaMA2-7B} & \textbf{LLaMA2-13B} & \textbf{LLaMA3-8B} \\
        \midrule
			{\methodname}   & 62.3 & 63.9 & 56.6 \\
			Post-Training Rotation  & 58.7 (\textcolor{myred}{$\downarrow$3.6}) & 61.3 (\textcolor{myred}{$\downarrow$2.6}) & 55.2 (\textcolor{myred}{$\downarrow$1.4}) \\
   \bottomrule
	\end{tabular}}
\end{table}   

\noindent\textbf{\underline{Where} to Apply Rotation?} In Figure~\ref{fig:rotation_overview}, we introduce two types of rotation in our pipeline, namely Between-Block Rotation applied on all weight matrices and In-Block Rotation applied on \textit{down\_proj} in FFN. As discussed in Section~\ref{sec:apply_rotation}, we can also apply a similar head-wise IBR $R_3$ for self-attention. The $R_3$ rotates the $W_v$ and $W_o$ in Figure~\ref{fig:rotation_overview} by $W^R_v \leftarrow W_v R_3, W^R_o \leftarrow R_3^{-1} W_o$.
 These choices for rotation targets are verified on LLaMA2-7B W4A4 PTQ shown in Table~\ref{tab:rotate_where_ablation}. The results suggest that applying and only applying both $R_1$ and $R_2$ is the best option to eliminate outliers.

\begin{table}[h]  
\caption{Ablation on \textbf{where} to apply rotation.}\label{tab:rotate_where_ablation}
\centering
\resizebox{0.48\textwidth}{!}{\begin{tabular}{lcc}
\toprule
\textbf{Method} & \textbf{Rotation} & \textbf{ZCSR$^7$ Avg.}  \\ 
\midrule
{\methodname} & $R_1, R_2$ & 54.1  \\
\midrule
\textcolor{myred}{\textbf{$(-)$}} FFN In-Block Rotation & $R_1$  & 40.4 (\textcolor{myred}{$\downarrow$13.7}) \\
\textcolor{myred}{\textbf{$(-)$}} Between-Block Rotation & $R_2$  & 49.7 (\textcolor{myred}{$\downarrow$4.4}) \\
\textcolor{myred}{\textbf{$(+)$}} Attention In-Block Rotation & $R_1,R_2,R_3$  & 53.8 (\textcolor{myred}{$\downarrow$0.3}) \\
\bottomrule
\end{tabular}}
\end{table} 

\noindent\textbf{\underline{How} to Apply LoRA?} In Section~\ref{sec:raf}, we propose two rotation-aware fine-tuning schemes LoRA After Rotation (LAR) and LoRA Before Rotation (LBR) shown in Figure~\ref{fig:rotation-lora-scheme}. We prove that LAR is the better paradigm based on the approximation error analysis compared with full-finetuning. In Table~\ref{tab:lora_how_ablation}, we quantitatively compare the W4A4 quantization performance of two schemes on the fine-tuning of the LLaMA2-7B. The LAR scheme demonstrates better effectiveness, which corresponds to the approximation analysis shown in Figure~\ref{fig:lora_approximation_rotate}.

\begin{table}[h]  
	\caption{Ablation on \textbf{how} to apply LoRA.}\label{tab:lora_how_ablation} 
	\centering
	\resizebox{0.49\textwidth}{!}{
		\begin{tabular}{cccc}
	\toprule
			\textbf{\#Bits-Quantizer} & \textbf{Method}  & \textbf{ZCSR$^7$ Avg.} & \textbf{MMLU$^4$ Avg.} \\
        \midrule
			\multirow{2}{*}{W4A4-GPTQ} & LAR  & 62.3 &  31.0 \\
			 & LBR  & 61.1 (\textcolor{myred}{$\downarrow$1.2}) & 30.4 (\textcolor{myred}{$\downarrow$0.6}) \\
   \bottomrule
	\end{tabular}} 
\end{table}  

\noindent\textbf{Outliers} Retaining the outlier-free characteristic during LLM fine-tuning is the most important motivation for {\methodname}. To quantitatively validate the effect of outlier elimination, we use kurtosis $\kappa = \frac{\sum_i^k(\mathbf{x}_i-\mu)^4}{\sigma^4 + \epsilon}$ of the activation to measure the outlier presence, where $\mu$ and $\sigma$ are respectively the empirical mean and standard deviation of activation distribution. Generally, a large kurtosis value indicates an activation distribution with heavy tails and a higher likelihood of outliers. We visualize the kurtosis dynamic during fine-tuning with LoRA and {\methodname} in Figure~\ref{fig:kurtosis}. In the early training epochs, the rotation effectively suppresses the activation outliers. The rotation-aware fine-tuning can retain this optimal property. After fine-tuning with {\methodname}, as shown in Figure~\ref{fig:kurtosis}, the kurtosis $\kappa$ across all layers is significantly reduced, which further gives rise to the low quantization error compared to the LoRA baseline. We also compare the activation distribution of {\methodname} against LoRA across layers in Figure~\ref{fig:full_activation} in the Appendix.

\noindent\textbf{LoRA rank settings} We explore the robustness of LoRA and {\methodname} towards various rank settings $r \in \{4,8,16,32,64\}$ when fine-tuning LLaMA2-7B and evaluated on zero-shot commonsense reasoning tasks. The optimal rank setting for {\methodname} and LoRA are 16 and 32, respectively. The lower optimal rank indicates the potential of our {\methodname} to save trainable parameters. Overall, {\methodname} consistently outperforms LoRA regardless of the rank setting, demonstrating its robustness.

\begin{figure}[h]
  \centering
    \includegraphics[width=0.9\columnwidth]{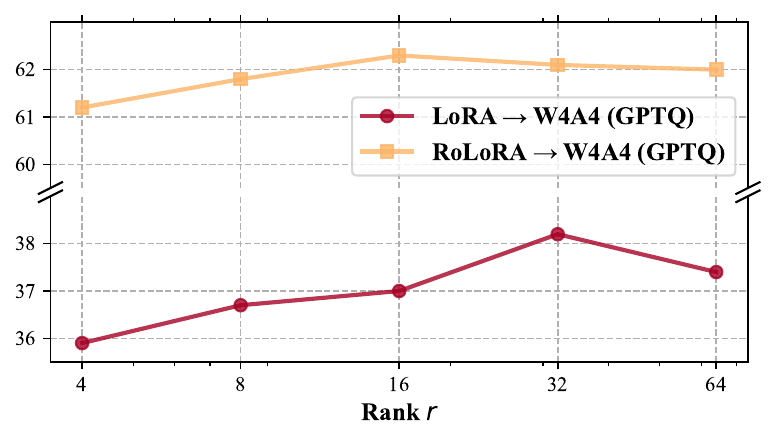} 
  \caption{Average accuracy of W4A4 LLaMA2-7B fine-tuned with {\methodname} for varying ranks $r$.}
  \label{fig:lora_rank}
\end{figure}

\noindent\textbf{Efficiency} For the fine-tuning efficiency of {\methodname}, the additional training time is only incurred by the online rotation operation ($R_2$ in Figure~\ref{fig:rotation_overview}) as the other rotation ($R_1$ in Figure~\ref{fig:rotation_overview}) can be directly merged into the original weights. There is only one additional matrix multiplication, and the increased rotation parameter can theoretically be considered negligible. We reported the fine-tuning cost of {\methodname} compared to LoRA in the same settings (rank $r=16$, batch size as 8, 3 total epochs) in Table~\ref{tab:training_cost}, where {\methodname} significantly improve W4A4 quantized LLaMA2-7B performance with extremely low additional overhead.

\begin{table}[h]  
	\caption{The fine-tuning costs comparison on LLaMA2-7B with batch size as 8 on NVIDIA H800 80G GPUs.} 
	\label{tab:training_cost}
	\centering
	\resizebox{0.48\textwidth}{!}{
		\begin{tabular}{ccccc}
			\toprule
			\textbf{Method}  & \textbf{Training Time} & \textbf{GPU Memory} & \textbf{ ZCSR$^7$ Avg.} \\
        \midrule
			LoRA            & 3.55 h & 23.0 GB & 37.0 (GPTQ) \\
			\rowcolor{gray!10} {\methodname}   & 3.65 h & 23.1 GB & 62.3 (GPTQ) \\
			\hline
	\end{tabular}}
\end{table}   

\section{Conclusion}
This paper presents {\methodname}, the first work to explore the feasibility of \textit{weight-activation} quantization in LoRA. {\methodname} applies rotation for eliminating outliers in activation distribution and performs rotation-aware fine-tuning to preserve the outlier-free characteristics. We theoretically and empirically investigate how to integrate rotation into LoRA. {\methodname} improves the performance of W4A4 and W6A6 LLMs by a great margin across various tasks with the same training cost. Moreover, {\methodname} can also help visual instruction tuning.


\section*{Limitation}
In this work, we propose a rotation-based fine-tuning method that can effectively improve quantization robustness to low-bit \textit{weight-activation} PTQ via retaining the outlier-free characteristics. The fine-tuning is conducted on NVIDIA H800 GPUs, while the recent NVIDIA Blackwell-architecture GPUs with 4-bit floating point support may further improve the efficiency. We will take the limitations into account and improve in future work. 

\section*{Acknowledgement}
This research is supported by HKSAR RGC General Research Fund (GRF) \#16208823.

\bibliography{acl_latex}

\newpage

\onecolumn
\appendix

\section{Detailed Evaluation Results}

Table~\ref{tab:zeroshot_full} and Table~\ref{tab:mmlu_full} listed the full evaluation results on zero-shot commonsense reasoning tasks and MMLU benchmarks, respectively. We use the `acc\_norm' in the evaluation report given by EleutherAI evaluation harness~\cite{gao2021framework} as the accuracy if there are such metrics. Otherwise, we use `acc'.

\begin{table*}[h]
\caption{Full accuracy comparison on zero-shot commonsense reasoning tasks of LLaMA series.}\label{tab:zeroshot_full}
\centering \small
\begin{tabular}{ccccccccccc}
\toprule
\textbf{\#Bits} & \textbf{Quantizer} & \textbf{Method} & \textbf{BoolQ} & \textbf{PIQA} & \textbf{HellaS.} & \textbf{WinoG.}  & \textbf{Arc-e} & \textbf{Arc-c} & \textbf{OBQA} & \textbf{Avg.} \\ 
\midrule
\multicolumn{11}{c}{LLaMA2-7B} \\
\midrule
FP16  & - & LoRA & 81.2 & 79.8 & 78.6 & 70.6 & 73.9 & 47.7 & 46.8 & 68.4  \\
\midrule
\multirow{4}{*}{W4A4}  & \multirow{2}{*}{RTN} & LoRA & 46.0 & 49.5 & 27.0 & 49.6 & 27.8 & 24.2 & 26.8 & 35.8 \\
&  & {\methodname} & 67.1 & 67.7 & 59.7 & 56.9 & 58.3 & 35.0 & 34.2 & 54.1  \\
\cmidrule{2-11}
& \multirow{2}{*}{GPTQ} & LoRA & 52.3 & 52.5 & 26.9 & 50.4 & 28.6 & 25.3 & 22.8 & 37.0 \\
&  & {\methodname} &  73.5 & 76.2 & 71.8 & 64.1 & 67.7 & 42.2 & 40.4 & 62.3  \\
\midrule
\multirow{4}{*}{W6A6}  & \multirow{2}{*}{RTN} & LoRA & 76.3 & 78.0 & 75.3 & 69.2 & 71.2 & 45.7 & 41.6 & 65.3 \\
&  & {\methodname} & 77.9 & 79.1 & 76.3 & 68.5 & 74.8 & 47.3 & 43.6 & 66.8  \\
\cmidrule{2-11}
& \multirow{2}{*}{GPTQ} & LoRA & 76.3 & 78.2 & 75.4 & 69.5 & 72.1 & 46.1 & 40.8 & 65.5  \\
&  & {\methodname} & 77.4 & 79.1 & 76.5 & 70.4 & 75.2 & 47.2 & 44.0 & 67.1  \\
\midrule
\multicolumn{11}{c}{LLaMA2-13B} \\
\midrule
FP16  & - & LoRA & 83.9 & 81.2 & 80.9 & 74.2 & 74.4 & 51.3 & 47.6 & 70.5 \\
\midrule
\multirow{4}{*}{W4A4}  & \multirow{2}{*}{RTN} & LoRA & 39.8 & 52.1 & 26.1 & 45.7 & 25.9 & 25.8 & 25.4 & 34.4 \\
&  & {\methodname} & 70.6 & 73.9 & 67.2 & 59.6 & 66.8 & 38.7 & 34.2 & 58.7 \\
\cmidrule{2-11}
& \multirow{2}{*}{GPTQ} & LoRA & 38.0 & 50.2 & 26.0 & 49.0 & 25.9 & 26.4 & 25.4 & 34.4 \\
&  & {\methodname} & 74.0 & 77.2 & 73.9 & 66.0 & 73.3 & 43.9 & 38.8 & 63.9\\
\midrule
\multirow{4}{*}{W6A6}  & \multirow{2}{*}{RTN} & LoRA & 80.8 & 78.1 & 77.8 & 70.3 & 73.0 & 49.2 & 42.2 & 67.3 \\
&  & {\methodname} & 80.3 & 78.8 & 78.0 & 71.1 & 77.6 & 49.6 & 43.2 & 68.4 \\
\cmidrule{2-11}
& \multirow{2}{*}{GPTQ} & LoRA & 81.9 & 79.2 & 78.5 & 69.3 & 74.3 & 51.5 & 41.2 & 68.0 \\
&  & {\methodname} & 80.6 & 79.3 & 78.1 & 72.5 & 77.4 & 49.4 & 44.0 & 68.8  \\
\midrule
\multicolumn{11}{c}{LLaMA3-8B} \\
\midrule
FP16  & - & LoRA & 64.6 & 82.4 & 81.4 & 75.1 & 81.8 & 56.5 & 48.0 & 70.0 \\
\midrule
\multirow{4}{*}{W4A4}  & \multirow{2}{*}{RTN} & LoRA &46.7 & 52.2 & 29.7 & 47.6 & 29.3 & 24.7 & 26.6 & 36.7 \\
&  & {\methodname} & 58.0 & 67.3 & 57.7 & 56.0 & 49.0 & 30.2 & 31.8 & 50.0 \\
\cmidrule{2-11}
& \multirow{2}{*}{GPTQ} & LoRA & 42.5 & 54.4 & 29.4 & 49.0 & 31.1 & 22.5 & 27.0 & 36.6\\
&  & {\methodname} & 63.2 & 71.1 & 66.7 & 60.2 & 60.3 & 38.2 & 36.8 & 56.6 \\
\midrule
\multirow{4}{*}{W6A6}  & \multirow{2}{*}{RTN} & LoRA & 75.5 & 78.3 & 77.4 & 70.8 & 76.4 & 51.2 & 44.0 & 67.7  \\
&  & {\methodname} & 78.6 & 79.5 & 76.7 & 71.1 & 77.6 & 49.8 & 40.8 & 67.8 \\
\cmidrule{2-11}
& \multirow{2}{*}{GPTQ} & LoRA & 77.9 & 78.3 & 77.9 & 71.3 & 75.2 & 50.5 & 43.2 & 67.8   \\
&  & {\methodname} & 78.1 & 79.3 & 76.8 & 71.9 & 76.7 & 50.9 & 42.8 & 68.1  \\
\bottomrule
\end{tabular}
\end{table*}

\begin{table*}[h]
\caption{Full accuracy on MMLU Benchmark of LLaMA series.}\label{tab:mmlu_full}
\centering \small
\begin{tabular}{cccccccc}
\toprule
\textbf{\#Bits} & \textbf{Quantizer} & \textbf{Method} & \textbf{Hums.} & \textbf{Other} & \textbf{Social} & \textbf{STEM} & \textbf{Avg.}  \\ 
\midrule
\multicolumn{8}{c}{LLaMA2-7B} \\
\midrule
FP16  & - & LoRA & 41.5 & 50.8 & 48.2 & 34.7 & 43.5 \\
\midrule
\multirow{4}{*}{W4A4}  & \multirow{2}{*}{RTN} & LoRA & 24.2 & 24.8 & 22.7 & 21.7 & 23.5  \\
&  & {\methodname} & 24.7 & 26.2 & 27.2 & 25.7 & 25.8\\
\cmidrule{2-8}
& \multirow{2}{*}{GPTQ} & LoRA & 24.3 & 24.5 & 23.0 & 22.0 & 23.5\\
&  & {\methodname} & 30.1 & 33.0 & 32.0 & 29.4 & 31.0 \\
\midrule
\multirow{4}{*}{W6A6}  & \multirow{2}{*}{RTN} & LoRA & 35.4 & 40.6 & 37.5 & 30.4 & 35.9 \\
&  & {\methodname} & 38.2 & 45.4 & 44.7 & 35.2 & 40.5 \\
\cmidrule{2-8}
& \multirow{2}{*}{GPTQ} & LoRA & 34.2 & 39.4 & 39.4 & 30.6 & 35.7 \\
&  & {\methodname} & 37.8 & 46.1 & 46.2 & 34.9 & 40.8 \\
\midrule
\multicolumn{8}{c}{LLaMA2-13B} \\
\midrule
FP16  & - & LoRA & 49.6 & 59.2 & 59.9 & 42.8 & 52.4 \\
\midrule
\multirow{4}{*}{W4A4}  & \multirow{2}{*}{RTN} & LoRA & 25.0 & 25.7 & 23.4 & 22.4 & 24.2 \\
&  & {\methodname} & 28.9 & 32.5 & 33.2 & 28.4 & 30.5\\
\cmidrule{2-8}
& \multirow{2}{*}{GPTQ} & LoRA & 25.5 & 24.2 & 24.1 & 23.4 & 24.4 \\
&  & {\methodname} & 37.7 & 42.3 & 43.7 & 32.7 & 38.9 \\
\midrule
\multirow{4}{*}{W6A6}  & \multirow{2}{*}{RTN} & LoRA & 44.3 & 52.8 & 55.0 & 38.6 & 47.3\\
&  & {\methodname} & 45.0 & 52.9 & 55.2 & 39.1 & 47.7 \\
\cmidrule{2-8}
& \multirow{2}{*}{GPTQ} & LoRA & 44.8 & 54.7 & 53.8 & 39.0 & 47.6 \\
&  & {\methodname} & 45.6 & 53.7 & 55.2 & 38.7 & 47.9 \\
\midrule
\multicolumn{8}{c}{LLaMA3-8B} \\
\midrule
FP16  & - & LoRA & 57.4 & 70.7 & 72.8 & 52.7 & 62.7 \\
\midrule
\multirow{4}{*}{W4A4}  & \multirow{2}{*}{RTN} & LoRA & 23.6 & 24.3 & 23.7 & 21.8 & 23.3 \\
&  & {\methodname} & 30.8 & 34.5 & 33.5 & 30.5 & 32.1  \\
\cmidrule{2-8}
& \multirow{2}{*}{GPTQ} & LoRA & 24.6 & 23.0 & 23.4 & 24.3 & 23.9 \\
&  & {\methodname} & 36.0 & 42.2 & 43.6 & 33.5 & 38.5  \\
\midrule
\multirow{4}{*}{W6A6}  & \multirow{2}{*}{RTN} & LoRA & 49.7 & 63.0 & 64.4 & 47.2 & 55.3  \\
&  & {\methodname} & 52.7 & 67.5 & 70.0 & 51.1 & 59.4 \\
\cmidrule{2-8}
& \multirow{2}{*}{GPTQ} & LoRA & 48.8 & 61.8 & 63.9 & 45.7 & 54.3  \\
&  & {\methodname} & 52.9 & 68.3 & 69.6 & 50.4 & 59.4 \\
\noalign{\vspace{0.2em}}
\bottomrule
\end{tabular}
\end{table*}

\section{Hyper-parameters for Reproduction}

In Table~\ref{tab:hyperpara}, we list the detailed hyper-parameters for reproducing {\methodname} and LoRA results. We do not apply searches on any hyperparameters for better accuracy, all the settings for the LLaMA series and LLaVA align with the default settings of \citet{zheng2024llamafactory} and \citet{liu2024visual}.

\begin{table}[h]
\centering
\caption{Detailed hyper-parameters for fine-tuning different LLMs and LMMs.}
\resizebox{0.9\textwidth}{!}{
\begin{tabular}{ccccc}
\toprule
Model            & LLaMA2-7B & LLaMA2-13B & LLaMA3-8B  & LLaVA-1.5-7B  \\ 
\midrule
Epoch                & 3            & 3             & 3         & 1 \\
Batch Size (Per GPU) & 8            & 4             & 8         & 2 \\
Gradient Accumulation& 1            & 2             & 1         & 64 \\
Warmup Ratio         & 0.01         & 0.01          & 0.01      & 0.03 \\
Optimizer            & AdamW        & AdamW         & AdamW     & AdamW\\
LoRA Rank $r$        & 16           & 16            & 16        & 128 \\
LoRA Dropout         & 0            & 0             & 0         & 0.05 \\
LoRA Target          & $W_q, W_v$ & $W_q, W_v$ & $W_q, W_v$ & $W_q, W_k, W_v, W_o, W_u, W_d, W_g$ \\
Learning Rate        & 1$e^{-4}$    & 1$e^{-4}$     & 1$e^{-4}$  & 2$e^{-4}$  \\
\bottomrule
\end{tabular}}
\label{tab:hyperpara}
\end{table}

\section{Activation Distribution Visualization}

\begin{figure*}[h]
  \centering
    \includegraphics[width=0.49\columnwidth]{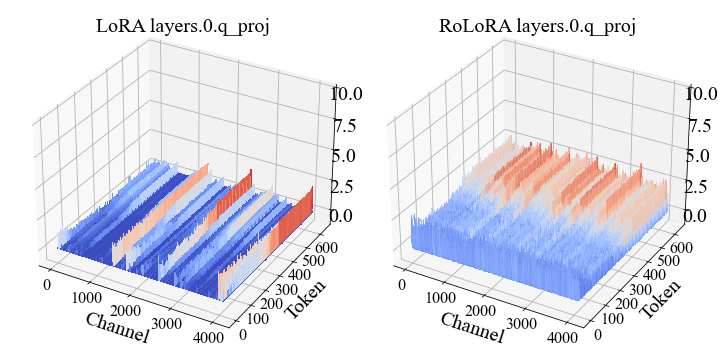} 
    \includegraphics[width=0.49\columnwidth]{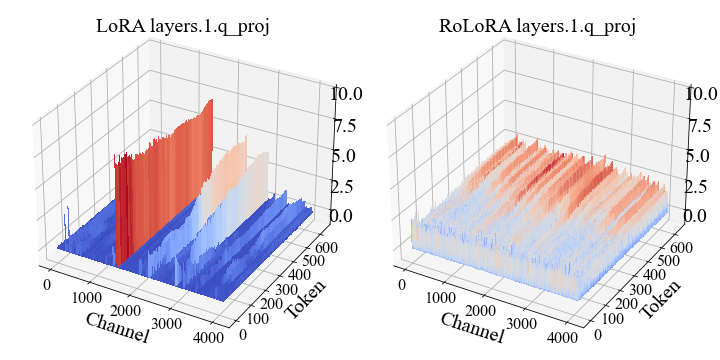} 
    \includegraphics[width=0.49\columnwidth]{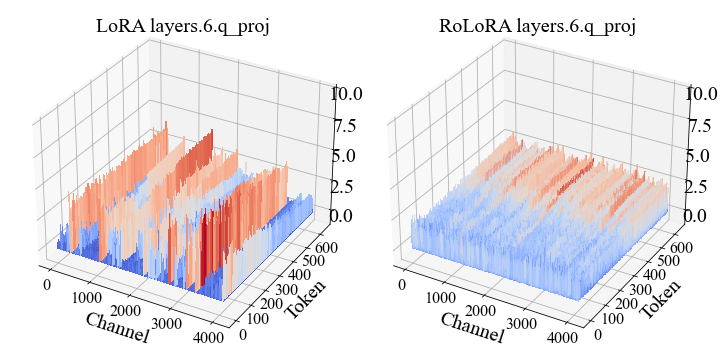} 
    \includegraphics[width=0.49\columnwidth]{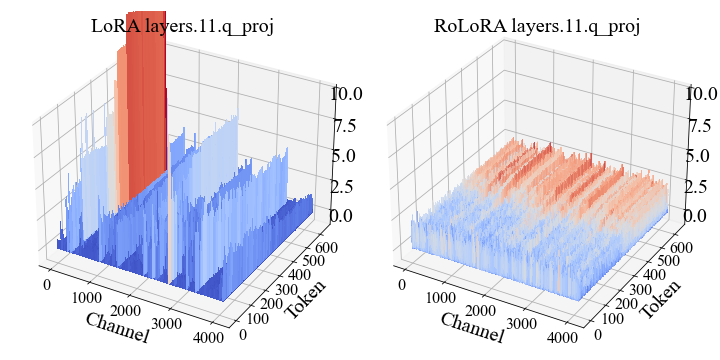} 
    \includegraphics[width=0.49\columnwidth]{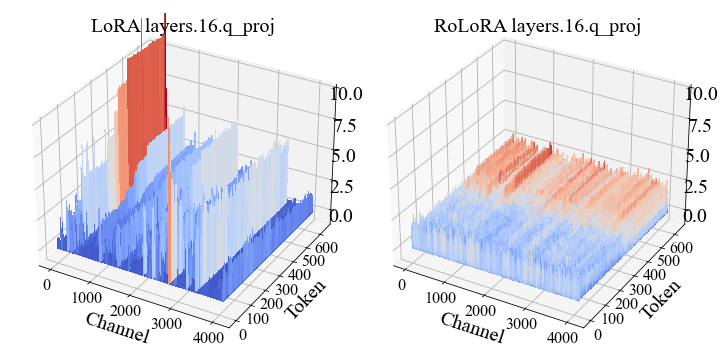} 
    \includegraphics[width=0.49\columnwidth]{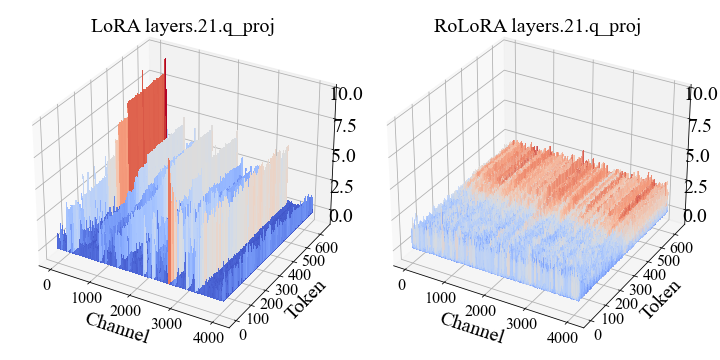} 
    \includegraphics[width=0.49\columnwidth]{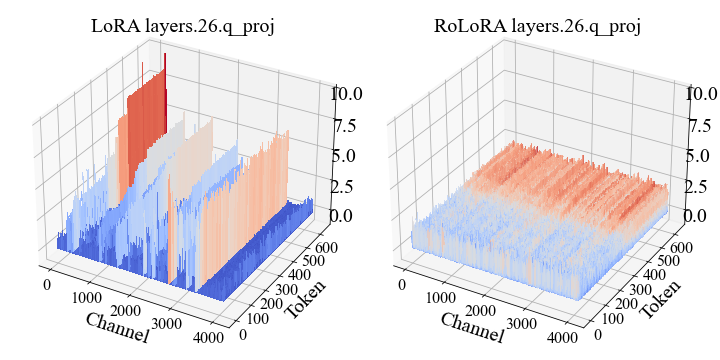} 
    \includegraphics[width=0.49\columnwidth]{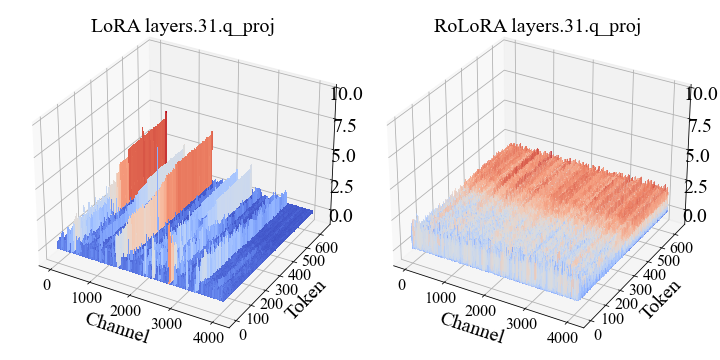} 
  \caption{Final activation distribution of the fine-tuned model produced using {\methodname} and LoRA. We select the output activation of \textit{q\_proj} across layers with the index of 0, 1, 6, 11, 16, 21, 26, 31.}
  \label{fig:full_activation}
\end{figure*}

We visualize the magnitude of the activation of fine-tuned LLaMA2-7B using LoRA and {\methodname} in Figure~\ref{fig:full_activation}. The visualizations reveal a noticeable amount of outliers presented
in the LoRA fine-tuned model, but are highly eliminated in {\methodname} counterpart.

\end{document}